\documentclass[letterpaper, 10 pt, conference]{ieeeconf}  % Comment this line out if you need a4paper

\IEEEoverridecommandlockouts                              % This command is only needed if 
\usepackage{graphicx} % for pdf, bitmapped graphics files
\usepackage{amsmath} % assumes amsmath package installed
\usepackage{amssymb}  % assumes amsmath package installed
\usepackage{booktabs}
\usepackage{subcaption}
\usepackage{url}
\usepackage{hyperref}
\usepackage{cite}
\usepackage{amsfonts}
\usepackage{algorithmic}
\usepackage{textcomp}
\usepackage{xcolor}
\def\BibTeX{{\rm B\kern-.05em{\sc i\kern-.025em b}\kern-.08em
    T\kern-.1667em\lower.7ex\hbox{E}\kern-.125emX}}
\begin{document}
\title{A Pairwise Human-Human Interaction Detection and Recognition Framework for Mobile Service Robots\\
\thanks{Sarah Gillet was partially supported by the Wallenberg AI, Autonomous Systems and Software Program – Humanity and Society (WASP-HS).}
}

\author{\authorblockN{Mengyu Liang}
\authorblockA{
% \textit{dept. Robot, Perception and Learning} \\
\textit{KTH Royal Institute of Technology}\\
Stockholm, Sweden \\
lmengyu@kth.se}
\and
\authorblockN{Iolanda Leite}
\authorblockA{
% \textit{dept. Robot, Perception and Learning} \\
\textit{KTH Royal Institute of Technology}\\
Stockholm, Sweden \\
iolanda@kth.se}
\and
\authorblockN{Sarah Gillet}
\authorblockA{
% \textit{dept. name of organization (of Aff.)} \\
\textit{MIT Media Lab}\\
Cambridge, MA, USA\\
sgillet@mit.edu}
% \and
% \authorblockN{4\textsuperscript{th} Given Name Surname}
% \authorblockA{\textit{dept. name of organization (of Aff.)} \\
% \and
% \authorblockN{5\textsuperscript{th} Given Name Surname}
% \authorblockA{\textit{dept. name of organization (of Aff.)} \\
% \and
% \authorblockN{6\textsuperscript{th} Given Name Surname}
% \authorblockA{\textit{dept. name of organization (of Aff.)} \\
}

\maketitle
% \thispagestyle{empty}
% \pagestyle{empty}

%%%%%%%%%%%%%%%%%%%%%%%%%%%%%%%%%%%%%%%%%%%%%%%%%%%%%%%%%%%%%%%%%%%%%%%%%%%%%%%%
\begin{abstract}

Autonomous mobile service robots, such as lawnmowers or cleaning robots, operating in human-populated environments need to reason about human-human interactions to support safe and socially aware navigation. For such systems, interaction understanding is not primarily a fine-grained recognition problem, but a perception problem under limited sensing quality and computational resources. Many existing approaches focus on holistic group activity recognition, often relying on complex and computationally expensive models that are not well suited for mobile robotic platforms.
In this work, we argue that pairwise human interactions constitute a minimal yet sufficient perceptual unit for robot-centric social understanding. We study the problem of identifying interacting person pairs and classifying coarse-grained interaction behaviors sufficient for downstream group-level reasoning and robot decision-making. To this end, we adopt a two-stage framework in which candidate interacting pairs are first identified using lightweight geometric and motion cues, and interaction types are subsequently classified using a relation network. We evaluate the proposed approach on the JRDB dataset, where it achieves competitive performance with reduced computational cost and model size compared to appearance-based methods. Additional experiments on the Collective Activity Dataset (CAD) and zero-shot evaluation on a lawnmower-collected dataset further demonstrate the generalizability of the proposed framework. These results suggest that simple geometric and motion cues provide a practical and efficient basis for interaction-aware perception in mobile service robots. Code can be found by \href{https://github.com/AsukaLmy/A-Pairwise-Human-Human-Interaction-Detection-and-Recognition-Framework-for-Mobile-Service-Robots}{this link}.

\end{abstract}

%%%%%%%%%%%%%%%%%%%%%%%%%%%%%%%%%%%%%%%%%%%%%%%%%%%%%%%%%%%%%%%%%%%%%%%%%%%%%%%%
\section{INTRODUCTION}
% start the motivation by robot
% Autonomous mobile robots operating in human-populated environments must continuously reason about nearby people to navigate safely and make appropriate decisions \cite{I_GCN_skeletonbased,Recent_trends_in_crowd_analysis:A_review,I_review_HOI}.
% For outdoor platforms such as delivery robots or lawnmowers, the primary requirement is not fine-grained action recognition, but a robust and coarse understanding of how nearby people are socially organized, for example, whether individuals are walking together, standing in a group, or sitting nearby. 
% Such interaction-level understanding provides actionable constraints for collision avoidance, path planning, and behavior modulation, while remaining compatible with the limited sensing quality and computational resources available on mobile robot.

Robots operating in human environments must continuously interpret social interactions to act safely and appropriately \cite{I_GCN_skeletonbased,Recent_trends_in_crowd_analysis:A_review,I_review_HOI}. For example, a service robot or an autonomous lawnmower navigating a public space needs to distinguish whether nearby people are moving independently or interacting as a group, as such interactions strongly influence motion prediction and collision avoidance. However, reliable perception of human interactions in real-world robotic settings remains challenging due to limited computational resources, noisy observations, and the need for real-time decision making.

Existing work has explored interaction and group behavior understanding from different perspectives. For instance, Ehsanpour et al. \cite{socialgroup_baseline1} demonstrated that complex models can effectively capture social structures and group behaviors, but such approaches are often computationally expensive and difficult to deploy in real-time robotic systems. To address this, Jahangard et al. \cite{JRDB_baseline1} propose a simplified model focusing on human relationship recognition. However, its scope is limited to relational structure and does not extend to explicit interaction behavior classification.

These limitations highlight a gap between expressive but heavy models and efficient but insufficient representations, motivating the need for a lightweight and behavior-aware interaction perception approach.

% This paragraph introduce the whole method brifly (can be seen as a summary of method and experiment)
To bridge this gap, we investigate whether minimal yet informative cues are sufficient for reliable interaction perception in real-world scenarios. We propose a lightweight two-stage framework that first identifies candidate interacting pairs using geometric cues, and then classifies interaction behaviors using a combination of motion, spatial, and visual features. The design emphasizes feasibility, efficiency, and robustness, making it suitable for deployment on resource-constrained robotic platforms.

To summarize, our contributions are as follows:

\begin{itemize}
    \item We introduce a robot centered formulation of human-human interaction recognition that focuses on detecting interacting pairs and coarse-grained interaction types, tailored to mobile service robotic applications.
    \item We propose an efficient two-stage pipeline that performs interaction detection and classification using bounding boxes and optical flow, avoiding reliance on skeleton-based representations.
    \item Through extensive ablation studies, we provide evidence that geometry and motion cues dominate performance for coarse-grained interaction classification, while complex appearance features offer limited additional benefit.
    We demonstrate that minimal interaction cues are sufficient to achieve reliable performance, highlighting a practical trade-off between model complexity and deployability for robotic systems.
    \item We evaluate our approach on the JRDB dataset and Collective Activity datasets, and demonstrate robust zero shot transfer to a mobile lawnmower platform, highlighting the practicality of the proposed interaction recognition method under real-world robotic conditions. %provide one of the first benchmarks for explicit pairwise interaction recognition in this setting.
\end{itemize}

\section{Related Work}
\subsection{Group Activity Recognition}
Group activity recognition (GAR) has been extensively studied due to its applications in surveillance, sports analysis, social scene understanding, and elderly care \cite{objd_Collective_Activity_Dataset,AR_Spatio-Temporal_Dynamic_Inference_Network_for_Group_Activity_Recognition,objd_first_one}. 
Most GAR methods adopt a holistic formulation that treats the scene as a single unit and predicts a dominant group-level activity~\cite{AR_Learning_Actor_Relation_Graphs_for_Group_Activity_Recognition, objd_Discriminative_Latent_Models_for_Recognizing_Contextual_Group_Activities,objd_Temporal_Model_for_Group_Activity_Recognition,objd_Social_Scene_Understanding}. 
While effective in offline analysis, such formulations, i.e., scene level activity recognitions, provide limited guidance for socially appropriate robot behavior. % be misaligned with the purpose of mobile robotic perception.
For example, when navigating through a public space, a robot may need to distinguish what individuals or subgroups of individual are doing, e.g., the robot needs to distinguish between walking individuals and multiple small sitting groups in order to choose a socially acceptable path.
From a robot’s perspective, only a subset of individuals are relevant but multiple interactions may happen within the same scene~\cite{I_GCN_skeletonbased, objd_regroup_T.A_2022,AR_Spatio-Temporal_Dynamic_Inference_Network_for_Group_Activity_Recognition}.
In crowded outdoor environments, accurately identifying whether individuals are interacting is more useful than recognizing a global group activity.
Rather than replacing group activity recognition, we view pairwise interaction modeling of two people as a more actionable intermediate representation for robots, which can be flexibly aggregated into group-level reasoning when needed.

% Group activity recognition (GAR) has been extensively studied in computer vision for understanding collective behaviors in surveillance and social scene analysis~\cite{objd_first_one}. Early approaches~\cite{objd_Discriminative_Latent_Models_for_Recognizing_Contextual_Group_Activities,objd_Temporal_Model_for_Group_Activity_Recognition,objd_Social_Scene_Understanding} relied on handcrafted features and spatio-temporal relationships among individuals, while more recent methods employ deep neural networks and graph-based formulations to model interactions implicitly across all people in a scene~\cite{objd_regroup_T.A_2022,AR_Learning_Actor_Relation_Graphs_for_Group_Activity_Recognition,Group_Spatio-Temporal-Dynamic-Inference-Network-for-Group-Activity-Recognition_2021}.
% Despite their success in predicting scene-level activity labels, most GAR methods adopt a holistic formulation, treating the entire scene as a single unit and inferring a dominant group activity. Such formulations are not well suited with robotic perception needs, as multiple local interactions may occur simultaneously in the robot's view. From those interactions, only a subset of nearby individuals is relevant for navigation and decision-making. 

\subsection{Pairwise Interaction Recognition}
Pairwise interaction recognition explicitly models the relationship between two individuals~\cite{I_GCN_skeletonbased,I_NTU_dataset, I_UT-Interaction-Data, I_TV_dataset,JRDB_baseline1}. % and offers a more fine-grained alternative to holistic group-level formulations.
By focusing on localized interactions, pairwise approaches are better suited to scenarios where multiple interactions coexist and explicit interaction localization is required~\cite{I_UT-Interaction-Data,I_TV_dataset}.
Existing pairwise approaches often rely on skeleton-based features ~\cite{I_GCN_skeletonbased}, which are fragile in outdoor scenarios due to occlusion, clutter, and low-resolution imagery, and often require computationally expensive processing pipelines~\cite{RN_Interaction_Relational_Network_for_Mutual_Action_Recognition}.

\subsubsection{Appearance- and Relation-based Approaches}
% \subsubsection{RGB-based Approaches}
To reduce dependence on pose quality, several works incorporate RGB appearance features into relational reasoning frameworks. Interaction relational networks~\cite{RN_Interaction_Relational_Network_for_Mutual_Action_Recognition} and attention-augmented relation models~\cite{RN_AARN}  combine visual features with relational modules to infer interactions between individuals, while non-local architectures~\cite{RN_non-local}  generalize relational reasoning to arbitrary region pairs. Although effective in controlled settings, such appearance approaches typically require deep visual backbones and high capacity relation modules, resulting in significant computational cost. For coarse-grained interaction recognition in robotic scenarios, such modeling method is often heavy to compute, especially when motion and spatial cues already provide strong discriminative signals. This raises questions about the accuracy–efficiency trade-off of appearance centered interaction models for robotic deployment.

\subsubsection{Geometry- and Motion-based Interaction Modeling}
Prior to the dominance of pose- and appearance-based methods, early work on interaction recognition explored handcrafted geometric and motion cues~\cite{AR_Learning_Actor_Relation_Graphs_for_Group_Activity_Recognition,AR_Spatio-Temporal_Dynamic_Inference_Network_for_Group_Activity_Recognition,group_groupformer_2021,Group_stagNet_2018}, such as relative distance, spatial overlap, and motion correlation between individuals. These approaches demonstrated that interaction patterns can be inferred from simple spatial and temporal relationships, with minimal computational requirements.
Our work revisits geometric and motion cues from a robot centered perspective as the core representation within a interaction detection and classification pipeline. By systematically integrating bounding box geometry and optical flow into a two-stage framework, we aim to retain the efficiency and robustness of cues while enabling explicit and pairwise interaction reasoning.

\section{Method}
\begin{figure*}[htbp]
    \centering
    \includegraphics[width=\textwidth]{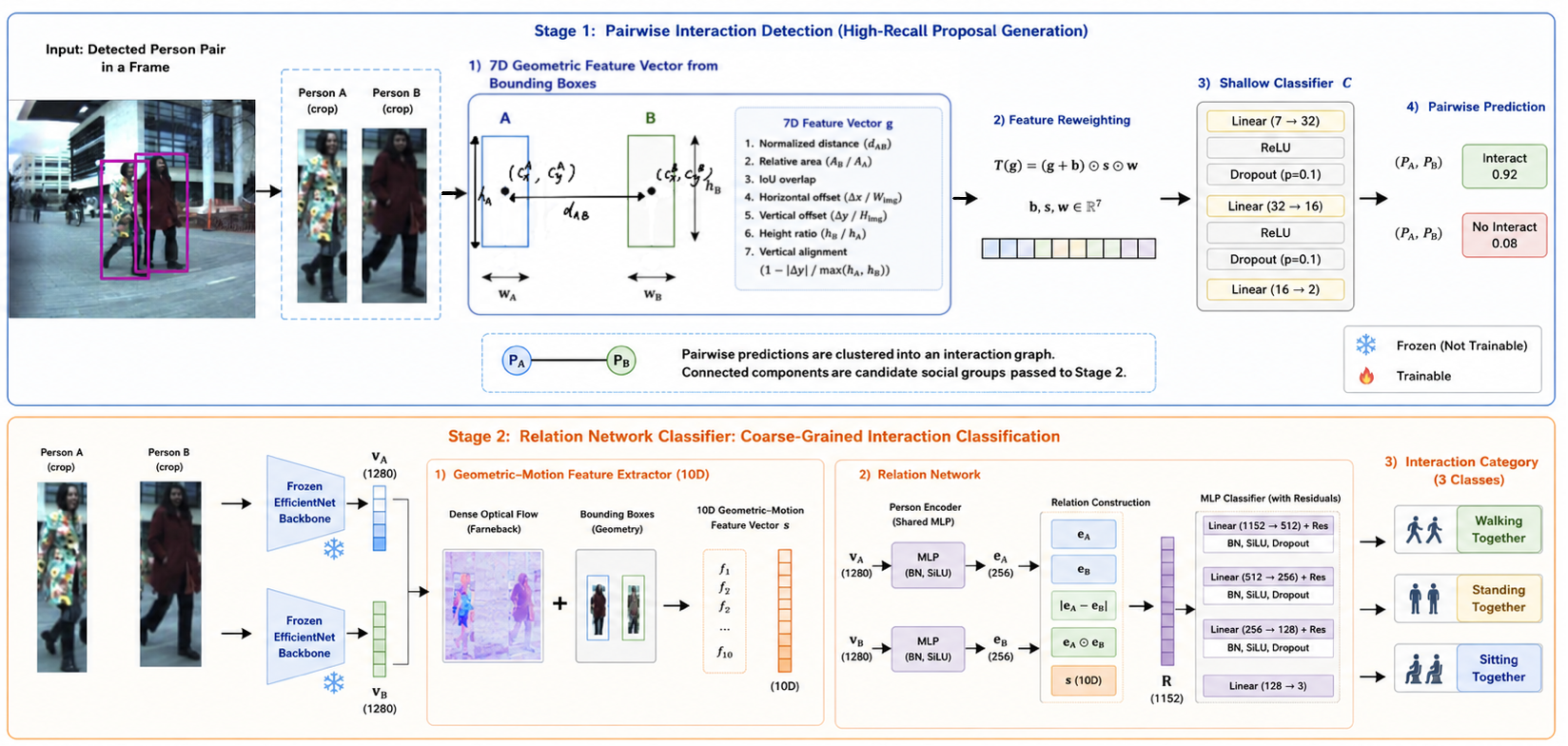}

    \caption{\textbf{Overview of the proposed two-stage pairwise interaction recognition framework:}\textbf{ Stage 1 }performs interaction detection using a $7D$ geometric feature vector derived from bounding box configurations, producing candidate interacting person pairs. \textbf{Stage 2 }classifies coarse-grained interaction types by combining frozen visual appearance features extracted by EfficientNet with geometric–motion features computed from optical flow, using a relation network for explicit pairwise reasoning. The framework is designed for efficient and robust deployment on constrained resource robotic platforms.}
    \label{fig:two_images}
\end{figure*}

%%%%%%%%%% stage2+1 %%%%%%%%%%%%%%%%%%%
%\subsection{Overall Architecture}
In this paper, we use interaction detection to refer to identifying whether a pair of individuals is interacting, interaction classification to denote categorizing the interaction type, and interaction recognition to describe the overall perception task combining both stages.

We adopt a two-stage framework for social interaction recognition, designed to balance interaction modeling capability with computational efficiency for mobile service robots. %robotic.
% The overall design explicitly prioritizes computational efficiency, data efficiency and deployment simplicity, making it suitable for resource-constrained mobile robotic platforms operating in outdoor environments.

\textbf{Stage 1} performs \textbf{pairwise interaction detection} as illustrated in Fig~\ref{fig:two_images}. We use geometric cues derived from people's bounding boxes to identify whether two individuals are socially interacting. This stage estimates whether two individuals are likely to be socially interacting. 
Rather than aiming for optimal interaction classification, Stage 1 serves as a high recall interaction proposal mechanism that filters out clearly non-interacting pairs for further analysis.
% This stage serves as an interaction proposal mechanism, filtering out non-interacting pairs and producing candidate interaction graphs.

\textbf{Stage 2} operates on the interaction proposals generated by Stage 1 and performs \textbf{coarse-grained interaction classification}. By restricting classification to a set of candidate interacting pairs, this stage integrates geometric and motion cues with frozen visual appearance features to refine interaction understanding while maintaining computational efficiency.

The overall architecture has the following components:

%The system consists of four main components:
\begin{itemize}
    \item \textbf{Geometric Interaction Module (Stage 1):} A classifier that infers pairwise interaction likelihoods from low-dimensional geometric features.
    \item \textbf{Frozen EfficientNet Backbone:} A pretrained EfficientNet~\cite{efficientnet} used for appearance feature extraction and kept frozen throughout training.
    \item \textbf{Geometric Flow Extractor:} A Farneback module~\cite{Farneback} that computes motion features between interacting persons.
    \item \textbf{Relation Network (Stage 2):} A trainable classification head that performs explicit relational reasoning over interacting person pairs.
\end{itemize}

The system predicts coarse-grained interaction categories that can be distinguished by motion or global pose changes sufficient for downstream robotic navigation and decision-making. In this work, we focus on \emph{Walking Together}, \emph{Standing Together}, and \emph{Sitting Together}.

\subsection{Feature Extraction Pipeline}

\subsubsection{Stage 1: Pairwise Interaction Detection}

Given detected person pairs within a frame, Stage 1 formulates interaction detection as a binary classification task. For each pair, we extract a $7D$ geometric feature vector from the corresponding bounding boxes, encoding relative distance, scale, overlap, and vertical alignment. These features provide a description of spatial relationships commonly associated with social interaction \cite{objd_regroup_T.A_2022}.

To account for the varying importance of different geometric cues, we apply a learnable feature reweighting transformation $T(\mathbf{g})$, followed by a shallow classifier $C$ to estimate pairwise interaction likelihoods:

\begin{equation}
T(\mathbf{g}) = (\mathbf{g} + \mathbf{b}) \odot \mathbf{s} \odot \mathbf{w}
\end{equation}

where $\mathbf{g} \in \mathbb{R}^{d_g}$ is the input geometric feature vector, $\mathbf{b} \in \mathbb{R}^{d_g}$ and $\mathbf{s} \in \mathbb{R}^{d_g}$ are learnable vectors and $\mathbf{w} \in \mathbb{R}^{d_g}$ is learnable weights. %All trainable parameters are initialized to one.
$\mathbf{s}$ and $\mathbf{w}$ are initialized to one, while $\mathbf{b}$ is initialized to zero to obtain a stable start.

We form classifier $C$ as 
\begin{multline}
C = \text{Linear}(d_g \rightarrow h_1) \rightarrow \text{ReLU} \rightarrow \\
\text{Dropout}(p) \rightarrow \text{Linear}(h_1 \rightarrow h_2) \rightarrow \\
\text{ReLU} \rightarrow \text{Dropout}(p) \rightarrow \text{Linear}(h_2 \rightarrow 2)
\end{multline}

with default configuration $h_1 = 32$, $h_2 = 16$, and dropout rate $p = 0.1$.

This design enables the model to adaptively emphasize informative geometric relationships while maintaining low computational cost.

At inference time, pairwise predictions can be clustered into an interaction graph where connected components represent candidate social groups for evaluation.  %through threshold $\theta$,
This stage is designed to prioritize recall over precision by limiting negative samples and using lower threshold, 
% $\theta$
reflecting the asymmetry of risk in robotic interaction perception: missing an interaction may have greater consequences for navigation and safety than temporarily considering a negative interaction.
The results are passed to Stage 2 for further classification.

\subsubsection{Stage 2: Appearance and Motion Features}

For each interacting person pair $p_i$ identified in Stage 1, Stage 2 extracts both appearance and motion cues before classifying coarse-grained interaction categories.

We obtain appearance features using a frozen EfficientNet \cite{efficientnet} backbone applied to cropped person regions. We opted for freezing the backbone to improve data efficiency and prevent overfitting on the limited interaction data we had available.

We extract optical flow field using dense optical flow computed between consecutive frames through the Farneback~\cite{Farneback} method. % We chose the Farneback method due to its computational efficiency and robustness. 
%We found that it is sufficient for capturing the coarse motion patterns relevant to mobile service robots.
%, e.g., walking, standing or .. together %to the target interaction categories.
From the optical flow field and bounding box geometry, we %compute a 10D geometric–motion feature vector 
encode relative distance, bounding box configuration, motion magnitude statistics, and motion consistency between two persons to compute a $10D$ geometric($f_1$ to $f_5$)-motion($f_6$ to $f_{10}$) feature. 

To quantify motion coordination between two people, we compute an interaction synchrony feature ($f_{10}$) that aggregates 4 similarity measures calculated from $f_3,f_6,f_7,f_8$, including flow magnitude similarity, temporal motion pattern similarity, posture consistency, and dominant motion direction alignment. The final geometric-motion feature vector is defined as:

\begin{equation}
\mathbf{g} =
[f_1, f_2, \dots, f_{10}]^T \in \mathbb{R}^{10}.
\end{equation}

To ensure permutation invariance with respect to person ordering, asymmetric features are computed from both $A \rightarrow B$ and $B \rightarrow A$ perspectives and symmetrized by averaging.

\subsection{Stage 2: Relation Network Classifier}

Stage 2 performs interaction classification using a relation network that explicitly models pairwise relationships.
We project each person’s appearance feature $v_A$, $v_B$ using a shared encoder 
\begin{equation}
\mathbf{e}_A = \text{PersonEnc}(\mathbf{v}_A), \quad \mathbf{e}_B =
\text{PersonEnc}(\mathbf{v}_B)
\end{equation}
where $\text{PersonEnc}$ is a shared two layer MLP with BatchNorm and SiLU activations.
  
The interaction representation is then constructed by combining the two person appearance feature with their encoded geometric–motion features $f$ using concatenation operations. Then the resulting interaction feature is passed to a MLP classifier to predict the interaction category.
\begin{multline}
  R = \text{Concat}(\mathbf{e}_A,\, \mathbf{e}_B,\, |\mathbf{e}_A -
  \mathbf{e}_B|,\, \mathbf{e}_A \odot \mathbf{e}_B,\, \mathbf{s}) \rightarrow \\
  \text{Linear}(1152 \rightarrow 512) \rightarrow \text{BN} \rightarrow \\
  \text{SiLU} \rightarrow \text{Dropout}(p) + \text{Res} \rightarrow \\
  \text{Linear}(512 \rightarrow 256) \rightarrow \text{BN} \rightarrow \\
  \text{SiLU} \rightarrow \text{Dropout}(p) + \text{Res} \rightarrow \\
  \text{Linear}(256 \rightarrow 128) \rightarrow \text{BN} \rightarrow \\
  \text{SiLU} \rightarrow \text{Dropout}(p) + \text{Res} \rightarrow \\
  \text{Linear}(128 \rightarrow 3)
\end{multline}

\section{Experiment}
We evaluate the proposed pairwise interaction method in terms of recognition performance, computational efficiency, and robustness across datasets and deployment scenarios relevant to mobile service robots.

\subsection{Dataset}
We evaluate our approach on two public benchmarks and one custom collected dataset. The public benchmarks provide annotations for human interactions and collective activities. Our dataset follows the same annotation scheme. % of the public benchmarks. 

JRDB~\cite{JRDB} is a dataset designed for human social understanding in human interactions across diverse indoor and outdoor social environments like cafeteria, teaching buildings and squares. %Recognizing humanity's innate social nature, 
By capturing videos of people in diverse environments, this dataset intends to offer an insight into human behavior within varied social contexts. For each video frame, the dataset provides annotations at three discernible levels: individual attributes, e.g., gender and age, intra-group interactions, e.g., standing and sitting together, and social group context, e.g., sitting together in cafeteria, making it well suited for evaluating pairwise interaction detection. %In this work, we only use ... % It provides robot-centric views with detailed annotations of human interactions, 

\begin{figure}
    \centering
    \includegraphics[width=1\linewidth]{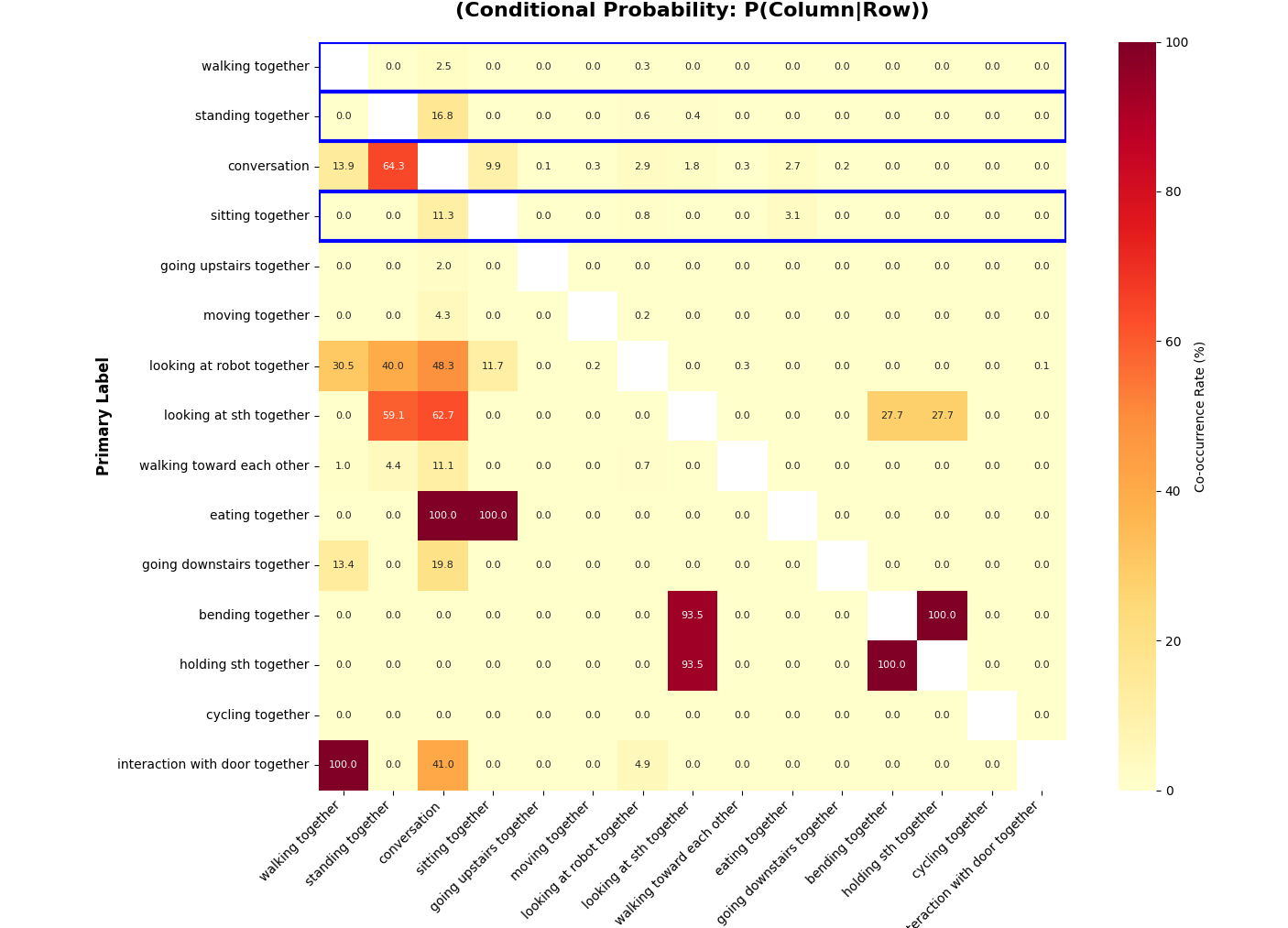}
    \caption{label distribution of JRDB, \textit{walking/standing/sitting together} collectively constitute 97.1\% of all interaction pairs and account for 88.6\% of the total label occurrences}
    \label{fig:labelJRDB}
\end{figure}

In this work, we only use the annotations for intra-group context. Further, we will focus on three categories: \textit{walking, standing and sitting together}. The reasons for only using these three categories are twofold: (1) these categories appear important for mobile robot social navigation (2) when inspecting the distribution of labels we found strong label dependencies. The label distribution is presented in figure~\ref{fig:labelJRDB}, \textit{conversation} frequently co-occurs with varying other behaviors, while \textit{eating together} only appears with \textit{conversation} and \textit{sitting together}. This indicates that additional labels are not independent modes but variations of the chosen three dominant categories. Therefore, reducing the label space to these three classes captures a majority of the dataset's essential structure which enables more robust modeling. We leave remaining labels and the evaluation of their importance for mobile robot navigation for future exploration. 

% For the interested reader, we provide  and their conditional probabilities in the Appendix. 

We additionally evaluate our framework on the Collective Activity Dataset (CAD)~\cite{objd_Collective_Activity_Dataset}, which contains videos of indoor and outdoor scenes with annotated group activities. Following prior work~\cite{socialgroup_baseline1}, we use the annotated social activity labels for training and evaluation. 

We also collected data using a forward-facing, low-angle camera mounted on a lawnmower at 6 scenes on a university campus, resulting in 14 minutes video streams with a resolution of 648×486 at 15 fps. The data is annotated with about 2k interactions in total for a zero shot evaluation.

\subsection{Data Filtering and Preprocessing}
%To better match the sensing conditions of mobile robotic platforms, 
We apply filtering to improve data reliability and remove false positive bounding boxes. We remove heavily occluded individuals and discard detections close to image boundaries, which were often incomplete, unstable or empty, i.e., did not resemble a person. Frames are sampled at an interval to reduce temporal redundancy while preserving interaction consistency.

During training, random horizontal flipping is applied for data augmentation. Class imbalance is handled through weighted sampling combined with a focal loss.

\subsection{Implementation Details}
We use EfficientNet \cite{efficientnet} as the default visual backbone due to its performance, and kept it frozen throughout training. To analyze the impact of appearance features, we additionally evaluate ResNet \cite{resnet}, VGG \cite{vgg}, AlexNet \cite{alexnet}, and MobileNet \cite{mobilenet} under the same training protocol.

The model is trained using AdamW \cite{AdamWoptimizer} with a fixed learning rate of $1\times10^{-4}$ and weight decay of $1\times10^{-5}$.

\begin{table*}[tbp]
\centering

\begin{tabular}{ccccccc}
\hline
\textbf{appearance} & \textbf{motion} & \textbf{geometric} & \textbf{Acc/\%} & \textbf{MPCA/\%} & \textbf{Macro F1} & \textbf{flops/Mb}\\
\hline
/ & / & y & $59.3\pm3.1$& $66.1\pm2.0$& $0.644\pm0.020$& 1.08\\
/ & y & / & $78.4\pm0.3$& $56.5\pm0.0$&$0.554\pm0.002$ & 1.08\\
/ & y & y & $84.3\pm1.8$& $79.6\pm0.9$&$0.772\pm0.008$ & 1.12\\
efficientnet\_v2\_s & y & y & $ 80.0\pm 2.6$ & $80.3\pm 1.1$ &$0.765\pm0.012$ & 5800\\
efficientnet\_v2\_s & / & y & $78.7\pm 2.4$& $78.4\pm1.8$& $0.774\pm0.015$& 5800\\
efficientnet\_v2\_s & y & / & $74.6\pm 3.5$&$70.8\pm2.3$ &$0.710 \pm0.015$& 5800\\
efficientnet\_v2\_s & / & / & $77.1\pm3.0 $& $69.6\pm1.5$& $0.711\pm0.012$& 5800\\
mobilenet\_v3\_s & y & y &75.5 &72.8 &0.727 & 122.51\\
resnet18 & y & y &74.8 & 76.3& 0.765& 975.16\\
vgg16 & y & y & 74.8& 74.6& 0.724& 7680\\
alexnet & y & y & 75.0&77.2 &0.710 &  292.26\\

\hline
\end{tabular}
\caption{Ablation study of visual, motion, and geometric cues on JRDB with Accuracy (Acc), Mean Per-Class Accuracy (MPCA), Macro F1-score, and computational cost (FLOPs). Result with $\pm$ are tested with three random seed: 42,59,108.
The results show that motion and geometric cues dominate performance, achieving the highest accuracy without relying on a visual backbone. Incorporating appearance features provides limited or negative gains while increasing computational cost, highlighting a favorable accuracy–efficiency trade-off for geometry and motion features in robotic perception.}
\label{tab:S2ablation}
\end{table*}

\begin{table}[htbp]

\centering
\begin{tabular}{l c c}
\toprule
& \textbf{Membership} & \textbf{Social Activity} \\
& Acc.\% & Acc.\% \\
\midrule
ARG \cite{AR_Learning_Actor_Relation_Graphs_for_Group_Activity_Recognition}[group] & 54.4 & 47.2 \\
I3D-SA-GAT \cite{socialgroup_baseline1}[group] & 54.4 & 47.7 \\
GT[group] & 54.4 & 51.6 \\
\addlinespace
ARG \cite{AR_Learning_Actor_Relation_Graphs_for_Group_Activity_Recognition}[individuals] & 62.4 & 49.0 \\
I3D-SA-GAT \cite{socialgroup_baseline1}[individuals] & 62.4 & 49.5 \\
GT[individuals] & 62.4 & 54.9 \\
\addlinespace
I3D-SA-GAT \cite{socialgroup_baseline1}[cluster] & 78.2 & 52.2 \\
I3D-SA-GAT \cite{socialgroup_baseline1}[learn2cluster] & 83.0 & 69.0 \\
ours & 79.8 & 52.6 \\ 
% 70.0 & 66.3/
\bottomrule
\end{tabular}
\caption{Comparison with state-of-the-art methods on the Collective Activity Dataset (CAD): Our pairwise interaction based approach is evaluated under both group-level and individual-level settings by mapping predicted interactions to group structures.
The results demonstrate that explicit modeling of pairwise interactions outperforms naive grouping baselines and achieves competitive performance compared to more complex clustering based and fully supervised methods, while maintaining lower model complexity.}
\label{tab:CAD results11111}
\end{table}

\subsection{Evaluation Metrics}

We report Accuracy, Mean Per-Class Accuracy (MPCA), and Macro F1-score to account for class imbalance:

\begin{equation}
\text{Accuracy} = \frac{\sum_{i=1}^{N} [\hat{y}_i = y_i]}{N}    
\end{equation}

\begin{equation}
\text{MPCA} = \frac{1}{C} \sum_{c=1}^{C} \frac{TP_c}{N_c} = \frac{1}{C}\sum_{c=1}^{C} \text{Recall}_c
\end{equation}

\begin{equation}
\text{Macro-F1} = \frac{1}{C} \sum_{c=1}^{C} \frac{2 \cdot P_c \cdot R_c}{P_c+R_c}
\end{equation}
where $P_c = TP_c / (TP_c + FP_c)$ and $R_c = TP_c / (TP_c + FN_c)$ are per-class precision and recall, respectively.

Computational efficiency is evaluated using FLOPs, particularly relevant for resource constrained robotic deployment.

\subsection{Ablation Study}
We conducted an ablation study to evaluate the contribution of appearance, motion, and geometric cues in the JRDB dataset. Results are presented in Table~\ref{tab:S2ablation}.

% The results show that motion and position features alone achieve the best overall performance, reaching the highest accuracy (0.843) and MPCA (0.801), without relying on any pre-trained visual backbone. Using motion features only yields competitive accuracy (0.784), while position-only features perform substantially worse, indicating that spatial cues alone are insufficient for reliable interaction recognition.
The results show that motion and geometric features dominate performance, achieving the best overall accuracy of $84.3\%$ without using a pretrained visual backbone. Motion features seem to have the strongest influence with a sole performance of $78.4\%$ accuracy. Geometry features reached a sole accuracy of $59.3\%$, indicating that spatial proximity alone is insufficient for reliable interaction recognition.

Incorporating visual appearance features does not seem to improve accuracy and often leads to degradation, suggesting redundancy with motion and geometric cues. EfficientNet yields slightly improved class balanced performance, but without clear gains in overall accuracy. This suggests improved class balance rather than stronger discriminative power. Focusing on efficiency, removing visual backbones reduces computational cost, suggesting a favorable accuracy–efficiency trade-off for robotic perception.

\subsection{Comparison to Baselines}
We compare our approach with state-of-the-art methods on CAD. Existing methods are primarily designed for group detection and group level activity recognition, whereas our approach focuses on pairwise interaction detection. The results are captured in Table \ref{tab:CAD results11111}.

To enable fair comparison, we map pairwise interaction predictions to group structures by connecting interacting individuals, and determine group activities via majority voting over predicted interaction types.

On the CAD dataset, our approach ($79.8\%$, $52.6\%$) consistently outperforms naive baselines that assume a single group ($54.4\%$, $51.6\%$) or no grouping ($62.4\%$, $54.9\%$) in grouping, reinforcing prior findings that highlight the importance of modeling group membership. Compared to unsupervised clustering methods ($78.2\%$, $52.2\%$), our approach achieves comparable or improved grouping accuracy while maintaining lower model complexity. Although fully supervised grouping methods ($83.0\%$, $69.0\%$) achieve higher accuracy, they require larger models and stronger supervision, making them less suitable for real-time robotic deployment.

\subsection{Zero Shot Inference on Lawnmower Dataset}
% \begin{figure*}[htbp]
%     \centering
    
%     \begin{subfigure}[b]{1\textwidth}
%         \centering
%         \includegraphics[width=\textwidth]{img/frame1_mosic.png}
%         \caption{}
%         \label{fig:1}
%     \end{subfigure}
    
%     \vspace{0cm}
    
%     \begin{subfigure}[b]{1\textwidth}
%         \centering
%         \includegraphics[width=\textwidth]{img/3_frame3785_comparison.png}
%         \caption{}
%         \label{fig:2}
%     \end{subfigure}

%     \vspace{0cm}
    
%     \begin{subfigure}[b]{1\textwidth}
%         \centering
%         \includegraphics[width=\textwidth]{img/frame4_m.png}
%         \caption{}
%         \label{fig:3}
%     \end{subfigure}
    
%     \caption{\textbf{Qualitative zero shot interaction recognition results on data collected from a mobile lawnmower platform}
%     (a) The lawnmower moves rapidly through a group of pedestrians, inducing strong ego-motion and viewpoint changes, which lead to misclassifications of interaction types.
%     (b) The lawnmower remains stationary, resulting in stable observations and correct interaction recognition most person pairs.
%     (c) The lawnmower approaches two individuals engaged in a face-to-face interaction, illustrating the intended use case of interaction aware perception to support downstream robotic decision-making.}
%     \label{fig:lawnmower}
% \end{figure*}

\begin{figure}[tbp]
    \centering
    \includegraphics[width=0.4\textwidth]{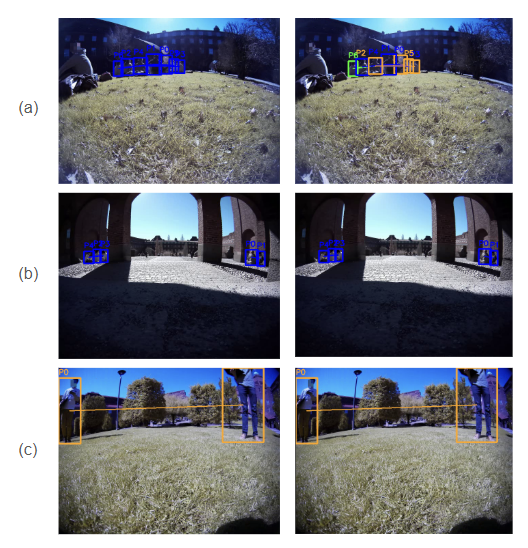}
    
    % \begin{subfigure}[b]{1\textwidth}
    %     \centering
    %     \includegraphics[width=\textwidth]{img/frame1_mosic.png}
    %     \caption{}
    %     \label{fig:1}
    % \end{subfigure}
    
    % \vspace{0cm}
    
    % \begin{subfigure}[b]{1\textwidth}
    %     \centering
    %     \includegraphics[width=\textwidth]{img/3_frame3785_comparison.png}
    %     \caption{}
    %     \label{fig:2}
    % \end{subfigure}

    % \vspace{0cm}
    
    % \begin{subfigure}[b]{1\textwidth}
    %     \centering
    %     \includegraphics[width=\textwidth]{img/frame4_m.png}
    %     \caption{}
    %     \label{fig:3}
    % \end{subfigure}
    
    \caption{\textbf{Qualitative zero shot interaction recognition results on data collected from a mobile lawnmower platform. (Complete results can be found \href{https://drive.google.com/file/d/1unozpsAtn7l7s9gSUNKOpLey4jNkahps/view?usp=sharing}{here}.)} Left column is prediction while right column is ground truth.
    (a) The lawnmower moves rapidly through a group of pedestrians, inducing strong ego-motion and viewpoint changes, which lead to misclassifications of interaction types.
    (b) The lawnmower remains stationary, resulting in stable observations and correct interaction recognition most person pairs.
    (c) The lawnmower approaches two individuals engaged in a face-to-face interaction, illustrating the intended use case of interaction aware perception to support downstream robotic decision-making.}
    \label{fig:lawnmower}
\end{figure}

We further evaluate our approach through zero-shot deployment on a mobile lawnmower platform operating in an outdoor environment, using the model trained on JRDB without any additional fine-tuning or retraining. Examples images and their annotations are shown in Figure~\ref{fig:lawnmower}.

% The data was collected using a forward-facing, low-angle camera mounted on the lawnmower at 6 scenes in school building, resulting in 14 minutes video streams with a resolution of 648×486 at 15 fps. The data is annotated with about 2k interactions in total.
% Person detections were obtained using YOLOv8, and the detected bounding boxes were used as input to our interaction perception pipeline.

Despite the domain shift in camera viewpoint, platform motion, and scene dynamics, our interaction detection stage achieves a precision of 96.5\%, successfully identifying the majority of interacting person pairs. 
Missed detections are primarily caused by low image resolution and partial visibility of distant pedestrians. %., rather than failures of the interaction model itself. 
Our method runs at 44 frames per second offline exceeding the camera frame rate of 15 fps, demonstrating the suitability of our method for real-time deployment on mobile robotic platforms.

For interaction type recognition, the stage 2 classifier achieves a macro F1-score of 0.51. 
Notably, the recall for the walking together and standing together classes reaches 98.5\% and 100\%, respectively, indicating strong robustness for common interaction patterns encountered during navigation. 
Most classification errors occur when sitting together interactions are misclassified as walking together. 
This behavior can be attributed to platform induced motion: although motion compensation is applied during optical flow computation, the rapidly changing visual content caused by lawnmower movement, including periodic wobbling due to irregularities in the grass surface, and low camera angle sometimes occluded by grass introduces flow patterns that are difficult to distinguish from human motion. 
Since the model is trained on data collected from static cameras, this represents a challenging zero shot transfer scenario.

\section{Discussion}
This work explores pairwise human interaction detection as a fundamental perceptual unit for robotic systems operating in outdoor environments. 
By explicitly focusing on local interactions rather than holistic group activity recognition and coarse-grained interaction categories, our approach emphasizes capturing what is most relevant for mobile robots, which typically require nearby and actionable social cues under limited sensing and computational resources.%aligns more closely with the needs of mobile robots, which typically require only nearby and actionable social cues for navigation and planning.

A key design choice of our method is the reliance on bounding box geometry and motion cues, avoiding skeletal representations and heavy visual backbones. The experimental results suggest that these lightweight cues are sufficient to capture essential interaction patterns for coarse-grained social activity recognition with similar performance or acceptable loss in accuracy. The lightweight nature of our approach offers advantages in computational efficiency and robustness. 
The ablation and baseline comparisons further indicate that motion and positional cues dominate performance, whereas increasingly complex visual features often yield diminishing returns.
% Taken together, these results highlight the importance of appropriate perceptual framework in robotic social understanding, and suggest that improving representation choices may be more effective than increasing model complexity for real-world service robot applications.
With these findings, we want to highlight that detailed appearance or pose information might not always be necessary for effective social activity recognition. We would like to encourage the community to complement the push for large models with the exploration of lightweight and simple approaches particularly when concerned with outdoor or resource constrained robotic settings.

\subsection{Limitations}
In the following, we would like to acknowledge limitations of the proposed approach. The coarse activity categories considered in this work may not capture more subtle or long-term social behaviors. 
In addition, the current framework assumes reliable person detection and tracking, and its performance may degrade under severe occlusions or in crowded scenes. Future work should further explore the suitability of this approach in varying scenarios and integrated into a navigating social robot.  % limited number of baselines due to training and evaluation on different dataset or code not availalble, but we used results from JRDB comparison paper where they showed that their approach was better than baselines

\section{Conclusion and Future Work}
This work presents pairwise interaction detection as a lightweight method for mobile service robots.
By relying on simple geometric and motion cues, the proposed approach captures interaction information that should be sufficient for socially-aware navigation, while avoiding the computational cost of complex visual or pose-based models.
These results suggest that appropriate representation choices, rather than increased model complexity, are practical to interaction perception on resource-constrained robotic platforms.

Future work may extend this framework by incorporating longer-term temporal reasoning, selectively integrating group context when necessary, or coupling interaction perception more tightly with downstream robotic planning and control.
Nonetheless, this study demonstrates that pairwise geometric and motion reasoning provides a strong and efficient foundation for interaction perception, offering a practical compromise between accuracy, interpretability, and computational cost for real-world mobile robotic systems.

%%%%%%%%%%%%%%%%%%%%%%%%%%%%%%%%%%%%%%%%%%%%%%%%%%%%%%%%%%%%%%%%%%%%%%%%%%%%%%%%

%%%%%%%%%%%%%%%%%%%%%%%%%%%%%%%%%%%%%%%%%%%%%%%%%%%%%%%%%%%%%%%%%%%%%%%%%%%%%%%%

%%%%%%%%%%%%%%%%%%%%%%%%%%%%%%%%%%%%%%%%%%%%%%%%%%%%%%%%%%%%%%%%%%%%%%%%%%%%%%%%

% \input{section/appendix}

\section*{ACKNOWLEDGMENT}

We would like to thank Malte Blomqvist for his support in running lawnmower experiments and data collection during the trials.

%%%%%%%%%%%%%%%%%%%%%%%%%%%%%%%%%%%%%%%%%%%%%%%%%%%%%%%%%%%%%%%%%%%%%%%%%%%%%%%%

% References are important to the reader; therefore, each citation must be complete and correct. If at all possible, references should be commonly available publications.

% \end{thebibliography}
\bibliographystyle{IEEEtran}
\bibliography{references}

\end{document}